\documentclass[11pt]{article}

\usepackage[margin = 1in]{geometry}

\usepackage{authblk}

\usepackage{microtype}
\usepackage{graphicx}
\usepackage{subfigure}
\usepackage{booktabs} 
\usepackage{natbib}
\usepackage{hyperref}
\usepackage{multicol}
\usepackage{multirow}  
\usepackage{colortbl}  
\usepackage{graphicx}
\usepackage{booktabs} 
\usepackage{pifont}   
\usepackage{xcolor} 

\usepackage{algorithm,algorithmic}

\usepackage{amsmath}
\usepackage{amssymb}
\usepackage{mathtools}
\usepackage{amsthm}

\usepackage{wrapfig}
\usepackage{threeparttable}

\usepackage[capitalize,noabbrev]{cleveref}

\theoremstyle{plain}

\theoremstyle{definition}

\theoremstyle{remark}

\usepackage[textsize=tiny]{todonotes}

\title{Training-Free Self-Correction for  Multimodal \\ Masked Diffusion Models}

\author[1]{Yidong~Ouyang}
\author[2]{Panwen~Hu}
\author[3]{Zhengyan~Wan}
\author[4]{Zhe~Wang}
\author[5]{Liyan~Xie}
\author[6]{Dmitriy~Bespalov}
\author[1]{Ying Nian Wu}
\author[1]{Guang~Cheng}
\author[7]{Hongyuan~Zha}
\author[2,8]{Qiang~Sun}

\affil[1]{\small 
  University of California, Los Angeles
}
\affil[2]{\small Mohamed bin Zayed University of Artificial Intelligence}
\affil[3]{\small 
East China Normal University
}
\affil[4]{\small 
University of Virginia
}

\affil[5]{\small
University of Minnesota}
\affil[6]{\small
Drexel university}
\affil[7]{\small The Chinese University of Hong Kong, Shenzhen}

\affil[8]{\small University of Toronto}

\date{}

\begin{document}

\maketitle

\vspace{-0.2in}

\begin{abstract}
Masked diffusion models have emerged as a powerful framework for text and multimodal generation. However, their sampling procedure updates multiple tokens simultaneously and treats generated tokens as immutable, which may lead to error accumulation when early mistakes cannot be revised. In this work, we revisit existing self-correction methods and identify limitations stemming from additional training requirements or reliance on misaligned likelihood estimates. We propose a training-free self-correction framework that exploits the inductive biases of pre-trained masked diffusion models. Without modifying model parameters or introducing auxiliary evaluators, our method significantly improves generation quality on text-to-image generation and multimodal understanding tasks with reduced sampling steps. Moreover, the proposed framework generalizes across different masked diffusion architectures, highlighting its robustness and practical applicability. Code can be found in \url{https://github.com/huge123/FreeCorrection}.
\end{abstract}

\section{Introduction}\label{sec:intro}
Masked diffusion models \citep{Austin2021StructuredDD, Hoogeboom2021ArgmaxFA, Sun2022ScorebasedCD, Lou2023DiscreteDM, Arriola2025BlockDI, Ou2024YourAD, Rutte2025GeneralizedID, Shi2024SimplifiedAG} have recently emerged as a powerful paradigm for text \citep{Nie2025LargeLD, Ye2025Dream7D, Cheng2025SDARAS} and multimodal generation \citep{Xin2025LuminaDiMOOAO, yang2025mmada}. By iteratively corrupting data through a masking process and learning to reverse this process, masked diffusion models enable parallel generation and offer an appealing alternative to autoregressive decoding. A growing body of work has shown that this framework scales well with model size and data diversity, and can be seamlessly integrated into large pre-trained architectures.

Despite their empirical success, masked diffusion models suffer from a fundamental limitation that is not yet fully solved. At each reverse step, masked diffusion models update a number of tokens simultaneously, and once a masked token is generated, it is typically treated as immutable in subsequent steps. As a result, errors introduced at early stages of generation may persist and accumulate, leading to mismatches between synthetic distribution and the target data distribution. An explanation of this phenomenon can be established from the perspective of continuous-time Markov chains (CTMCs) \cite{campbell2022continuous}. The practical reverse sampling procedure employed by masked diffusion models is mathematically equivalent to a $\tau$-leaping approximation of the underlying reverse-time CTMC \cite{Gillespie2001ApproximateAS}. This approximation marginalizes out exact jump times and performs parallel, approximate updates across all masked dimensions. While $\tau$-leaping enables efficient and highly parallel generation, it is well known in the numerical analysis of CTMCs to introduce path-wise inaccuracies when incorrect jumps occur early and cannot be revised. 

In this paper, we re-examine existing approaches that aim to mitigate generation errors via self-correction. Prior work typically relies on either (i) fine-tuning or retraining additional models to explicitly evaluate or re-score generated tokens \citep{Rutte2025GeneralizedID, Zhao2024InformedCF,Kim2025FineTuningMD, Peng2025PathPF, Huang2025DontST}, or (ii) reusing likelihood estimates obtained several steps earlier in the reverse process \citep{Wang2025RemaskingDD}. While effective in some settings, these approaches often introduce additional training costs and rely on misaligned signals. In contrast, our goal is to develop a training-free correction mechanism that operates directly on the reverse process and remains faithful to the underlying generative dynamics.

Our key insight is that masked diffusion networks already possess useful inductive biases that can be exploited for self-correction without additional training. By carefully analyzing the structure of the reverse process and the model’s conditional predictions, we identify more fine-grained design choices that allow the model to reassess and refine uncertain updates during generation. Importantly, our method does not require modifying the pre-trained model or introducing auxiliary evaluators. 

We validate our approach through GenEval benchmark \citep{Ghosh2023GenEvalAO} on text-to-image generation and VLMEvalKit \citep{Duan2024VLMEvalKitAO} on several multimodal understanding benchmarks. To the best of our knowledge, this is the first work to study self-correction mechanisms in multimodal masked diffusion models. Our results show that the proposed method significantly improves generation fidelity and semantic alignment, while simultaneously enabling faster sampling without sacrificing output quality. Moreover, the approach generalizes across different masked diffusion architectures, highlighting its robustness and practical applicability. Together with detailed ablation studies on the design of remasking criteria, our findings suggest that using the inductive bias of pre-trained masked diffusion models for remasking opens new avenues for efficient and principled self-correction at inference time.

\section{Related Work}
\paragraph{Masked Diffusion Models.}
Masked diffusion models (MDMs) generalize the denoising diffusion paradigm to discrete token spaces by corrupting data with a masking forward process and learning the corresponding reverse dynamics.
Early work studied discrete diffusion for discrete variables and established practical training and sampling objectives for masked corruption and parallel decoding~\citep{Austin2021StructuredDD,Hoogeboom2021ArgmaxFA,Sun2022ScorebasedCD,Lou2023DiscreteDM,Shi2024SimplifiedAG}.
Recent advances have focused on scaling masked diffusion to large language model settings and improving decoding efficiency/quality via better schedules and token-selection rules such as Top-$K$ updates~\citep{Nie2025LargeLD,Ye2025Dream7D,Cheng2025SDARAS}.

Beyond pure text, MDMs have also been extended to multimodal generation and unified vision-language modeling by first adopting VQ-VAE \citep{Oord2017NeuralDR} to convert images to discrete tokens, and then operating over image and text tokens through large-scale pretraining~\citep{Xin2025LuminaDiMOOAO,yang2025mmada}.
These models inherit the key advantage of diffusion-style parallel generation, but also amplify a known issue of discrete parallel sampling: once tokens are filled, they are typically treated as absorbing states, making early mistakes hard to revise.

\paragraph{Self-Correction of Masked Diffusion Models.}
A growing line of work investigates how to endow MDMs with self-correction so that previously generated tokens can be revised.
From a continuous-time Markov chain (CTMC) perspective, several methods adapt predictor-corrector style sampling to reduce discretization errors in the reverse dynamics~\citep{campbell2022continuous,Lezama2023DiscretePD,Gat2024DiscreteFM,campbell2024generative,Park2024JumpYS}.
Another family of approaches introduces explicit remasking in the reverse process: tokens can be re-masked according to a schedule and then re-generated, thereby enabling correction during sampling~\citep{Wang2025RemaskingDD,Peng2025PathPF}.

Existing self-correction methods mainly differ in whether they require additional training.
Some approaches generalize the diffusion process and train new models to support correction and improved reverse transitions~\citep{Rutte2025GeneralizedID,Peng2025PathPF}.
Others estimate per-token quality/confidence with auxiliary networks or by fine-tuning, and then use these estimates to decide which tokens to remask or re-sample~\citep{Zhao2024InformedCF,Kim2025FineTuningMD,Huang2025DontST}.
In contrast, our goal is to achieve \,\emph{training-free} self-correction for \emph{multimodal} MDMs by exploiting the inductive bias already present in pre-trained masked diffusion backbones, without modifying model parameters or introducing external evaluators.

\section{Preliminaries} 
\subsection{Formulation of Masked Diffusion Model}\label{sec:pre_mdm}
Masked diffusion models \citep{Austin2021StructuredDD, Hoogeboom2021ArgmaxFA, Sun2022ScorebasedCD, Lou2023DiscreteDM, Arriola2025BlockDI, Ou2024YourAD, Rutte2025GeneralizedID, Shi2024SimplifiedAG, Nie2025LargeLD} are characterized by the forward and backward processes. In the forward process, we gradually replace each word in a sentence with a special mask token until all the words are replaced. In the backward process, the model learns to gradually complete the sentence from all mask tokens.

In this work, we follow the Continuous-Time Markov Chain (CTMC) framework used in \citep{campbell2022continuous, Gat2024DiscreteFM, campbell2024generative, Liu2024ThinkWY} to model the masked diffusion models. We consider a $\mathcal{D}$-dimensional discrete state space $\mathcal{S}^{\mathcal{D}}$, where $\mathcal{S}$ denotes the finite state space of each dimension. 
The data distribution is denoted by $q_0$, with probability mass function $q_0(x)$ over $\mathcal{S}^{\mathcal{D}}$, and a data sample is written as $x_0 \in \mathcal{S}^{\mathcal{D}}$.


\paragraph{Forward Process.} For a given noise level $t \in[0,1]$, the forward process $x_t \sim q_{t| 0}\left(x_t | x_0\right)$ is a coordinate-independent masking process via $q_{t|0}(x_t|x_0)=\prod_{d=1}^{\mathcal{D}}q^d_{t|0}(x_t^d|x_0^d)$, where
\begin{equation}\label{eq:forward}
q_{t|0}^d(x_t^d|x_0^d)=(1-\alpha_t)\delta_{m}(x_t^d) +\alpha_t\delta_{x_0^d}(x_t^d), \, d \in [\mathcal D],     
\end{equation}
here $\alpha_t$ is the predefined noise schedule satisfying $\alpha_0=1, \alpha_1 =0$; $\delta_{x}(z)$ denotes Kronecker delta satisfying $\delta_{x}(z)=1$ if $x=z$ and $\delta_{x}(z)=0$ if $x\neq z$; and $m$ denotes the one-hot embedding of mask. In other words, for each coordinate $i$, at forward time step $t$, $x_t^i$ is replaced by the mask token with probability $1-\alpha_t$ and remains unchanged otherwise \cite{campbell2022continuous}.

\paragraph{Reverse Process.} The oracle reverse process of the above forward process, conditional on the initial state $x_0$, is denoted as $q_{s | t}\left(x_s | x_t, x_0\right)=\prod_{d=0}^{D} q_{s | t}\left(x_s^d | x_t, x_0\right)$ for any $s<t$, where
\begin{equation}\label{eq:reverse_process}
\begin{aligned}
 q_{s | t}\left(x_s^d | x_t, x_0\right)  = \begin{cases} \delta_{x_t^d}(x_s^d), & \text{ if }x_t^d \neq m, \\ 
\frac{1-\alpha_s}{1-\alpha_t} \delta_{m}(x_s^d) + \frac{\alpha_s-\alpha_t}{1-\alpha_t} \delta_{x_0^d}(x_s^d),
& \text{ if }x_t^d=m.\end{cases}    
\end{aligned}
\end{equation}

\paragraph{Training Objective.}
A neural network $p_\theta$ parametrized by $\theta$ can be trained to approximate the oracle reverse process by minimizing the Kullback–Leibler (KL) divergence between the true reverse conditional distribution $q_{s | t}\left(x_s^d | x_t, x_0\right)$ and $q_{s | t}\left(x_s^d | x_t, p_\theta^d\left(x_t, t\right)\right)$, where $p_\theta^d\left(x_t, t\right)$ denotes the $d$-th coordinate of the network output. The training objective can be summarized as
$$
\mathcal{L}_\theta=\int_0^1 \frac{\alpha_t^{\prime}}{1-\alpha_t} \underset{\substack{x_0 \sim q_0 \\ x_t \sim q_{t | 0}\left(x_t | x_0\right)}}{\mathbb{E}} \left[ 
 \sum_{d=1}^D\delta_{m}(x_t^d) x{_0^d}^{\top} \log p_\theta^d\left(x_t, t\right)\right] dt.
$$


Here, $\alpha_t^{\prime}=\frac{d \alpha_t}{d t}$. In practice, a time-embedding-free architecture for the denoising network, i.e., $p_\theta\left(x_t, t\right)=p_\theta\left(x_t\right)$, is usually employed as $x_t$ implicitly contains information about $t$ via the number of masked tokens.

\paragraph{Sampling Process.}
After learning the network $p_\theta(x_t)$, samples can be generated by simulating the reverse process in Eq.~\eqref{eq:reverse_process}.
Since the reverse dynamics factorize across dimensions, sampling can be performed independently for each coordinate.
Specifically, for each masked position $d$ at time $t$, a transition from the mask token occurs during the interval $(s,t]$ with probability
$\frac{\alpha_s-\alpha_t}{1-\alpha_t}$; conditioned on such a transition, the new token is sampled from the categorical distribution
$p_\theta^d(x_t)$.

\subsection{The Limitation of Masked Diffusion Model}
\label{sec:pre-limitation}

To understand the limitations of the sampling process in masked diffusion models better, we present a detailed discussion on the coordinate-wise independent, time-inhomogeneous CTMC below. 

First of all, it can be observed that, for each dimension $d$, the forward masking process admits the following time-varying transition rate:
\[
Q_t^{(f)}(x_0^d \to m) = \lambda(t):= -\frac{\alpha_t'}{\alpha_t},
\] 
since the resulting transition probability satisfies
\[
q(x_t^d = x_0^d|x_0^d) 
= \exp\!\left(-\int_0^t \lambda(\tau)\,d\tau\right)
= \alpha_t,
\]
which recovers the predefined noise schedule in Eq.~\eqref{eq:forward}.

The reverse process corresponds to the time-reversed CTMC, which progressively transforms the masked state back into samples from the data distribution. The reverse-time transition rate can be computed in two equivalent ways. 
One approach, adopted in \citep{campbell2022continuous,lou2023discrete}, expresses the reverse rate matrix $Q_t^{(r)}$ in terms of the forward rate as,
\begin{equation}\label{eq:reverse_rate_ratio}
\begin{aligned}
Q_t^{(r)}(x_t \to x_s)
= Q_t^{(f)}(x_s \to x_t)
\sum_{x_0}
\frac{q_{t | 0}(x_s | x_0)}
     {q_{t | 0}(x_t | x_0)}
\, q_{0 \mid t}(x_0 | x_t),
 x_t \neq x_s.
\end{aligned}
\end{equation}

Alternatively, following \citep{Gat2024DiscreteFM,campbell2024generative}, 
the reverse transition rate can be obtained via marginalization over the posterior of the data state:
\[
Q_t^{(r)}(x_t \to x_s)= \mathbb{E}_{q_{0 | t}(x_0 | x_t)}
\bigl[
Q_t^{(r)}(x_t \to x_s | x_0)
\bigr],
\]
where the posterior of the data state is given by
$
q_{0 \mid t}(x_0 \mid x_t)
= \frac{q_{t | 0}(x_t | x_0)\, q_0(x_0)}{\mathbb{E}_{q_0(x_0)}\!\left[q_{t | 0}(x_t | x_0)\right]}$, 
and \(Q_t^{(r)}(x_t \to x_s |x_0)\) denotes the conditional transition rate of the reverse-time CTMC given the initial state \(x_0\).
After the calculation, we get the reverse transition rate:
\[
Q_t^{(r)}(x_t \to x_s) =\frac{\alpha_t'}{1-\alpha_t} \delta_m(x_t) p_{0 | t}\left(x_0=x_s | x_t\right).
\]

The reverse CTMC could be simulated exactly using Gillespie’s Algorithm \citep{Gillespie1976AGM, Gillespie1977ExactSS, Wilkinson2006StochasticMF}, which samples the next jump time from an exponential distribution and selects a single dimension to update. As pointed out by \citep{campbell2022continuous}, this is inefficient for large $D$. In practice, various methods \citep{campbell2022continuous, lou2023discrete} adopt $\tau$-Leaping \citep{Gillespie2001ApproximateAS,Wilkinson2006StochasticMF} as a parallel algorithm to update all transitions in a small time interval simultaneously. The sampling process described in Section \ref{sec:pre_mdm} is a first-order approximation with $\tau$-Leaping in sampling of CTMC shown as follows.

The exact transition probability over a finite interval $[s,t]$ is given by integrating the transition rate. 
Since the CTMC is absorbing once unmasked, only transitions from $m$ to a data token $v$ are nontrivial. 
Denoting $x_t^d = m$, the transition probability satisfies the following Kolmogorov forward equation for any $\tau \in [s,t]$,
\[
\frac{d}{d\tau} q_{\tau | s}(v | m) = Q_\tau^{(r)}(m \to v) 
= \frac{\alpha_\tau'}{1-\alpha_\tau} \, p_{0 | \tau}(v | m).
\]
The transition probability thus can be written as
\begin{align*}
q_{s | t}(x_s^d = v \mid x_t^d = m)
&= \int_s^t \frac{\alpha_\tau'}{1-\alpha_\tau} \, p_{0 | \tau}(v \mid m) \, d\tau \\
&\approx p_{0|t}(v \mid x_t, t) 
   \int_s^t \frac{\alpha_\tau'}{1-\alpha_\tau} \, d\tau \\
&= p_{0|t}(v \mid x_t, t) \, \log \frac{1-\alpha_t}{1-\alpha_s} \\
&\approx p_{0|t}(v \mid x_t, t) \frac{\alpha_s - \alpha_t}{1-\alpha_t}.
\end{align*}

Therefore, we show that the sampling process of the masked diffusion model is a first-order approximation of $\tau$-leaping sampling of the underlying reverse-time CTMC.

A characteristic property of $\tau$-leaping is that approximate jumps across multiple dimensions are executed simultaneously. 
Due to the special design in the masked diffusion model, once a coordinate transitions from the mask token to a data token, it becomes absorbing. Consequently, errors introduced in early parallel jumps cannot be corrected in later steps. To mitigate the error accumulation caused by parallel generation, two main lines of work have been proposed.

The first line of work aims to restrict generation to tokens for which the model has high confidence.
In particular, the Top-K strategy \citep{Zheng2023ARD,Ye2024BeyondAD,Wang2024DiffusionLM,nie2025large,Wang2025RevolutionizingRL,Xin2025LuminaDiMOOAO} is widely adopted in modern masked diffusion models.
Instead of treating each position as being generated with probability $(\alpha_s-\alpha_t)/(1-\alpha_t)$, Top-K decoding selects the positions to be updated based on the highest predicted token probabilities. Top-K margin strategy \citep{Kim2025TrainFT} adopts the difference between the top two predicted probabilities. However, this type of method can only alleviate the error caused by parallel generation. 

Another line of work endows masked diffusion models with self-correction capabilities  \citep{campbell2022continuous, Lezama2023DiscretePD, Gat2024DiscreteFM, campbell2024generative, Wang2025RemaskingDD, Peng2025PathPF, Rutte2025GeneralizedID, Zhao2024InformedCF, Kim2025FineTuningMD, Park2024JumpYS}, allowing previously generated tokens to be revised in later steps. 

\subsection{Self-Correction of Masked Diffusion Models}
\citep{campbell2022continuous, Lezama2023DiscretePD, Gat2024DiscreteFM, campbell2024generative} adopt predictor-corrector for CTMC. 
\citep{Wang2025RemaskingDD, Peng2025PathPF} generalize the reverse transition probability such that a generated token can be remasked according to the schedule $\sigma_t$, while \citep{Rutte2025GeneralizedID} generalizes both forward and reverse processes for self-correction, which requires training a new masked diffusion model. \citep{Zhao2024InformedCF, Kim2025FineTuningMD, Huang2025DontST} define per-token quality as the criterion for remasking, and finetune a pre-trained model for estimation per-token quality. We summarize existing self-correction methods according to whether they require finetuning an existing model or training a new model in Table \ref{tab:method_category}.


\begin{table}[t]
\centering
\caption{Categorization of self-correction for masked diffusion models.}
\label{tab:method_category}
\small
\begin{tabular}{ll} 
\toprule
\textbf{Category} & \textbf{References} \\
\midrule
Training-free & \cite{Wang2025RemaskingDD} \\
Training-based or Finetuning-based & \cite{Rutte2025GeneralizedID, Zhao2024InformedCF, Kim2025FineTuningMD, Peng2025PathPF, Huang2025DontST} \\
\bottomrule
\end{tabular}
\end{table}

\begin{table*}[h]
\centering
\caption{Evaluation of text-to-image generation on the GenEval benchmark.
}
\label{tab:geneval}
\begin{threeparttable}
\begin{tabular}{l c c c c c c c}
\toprule
Method & Single & Two & Count & Color & Pos. & Attr. & Overall $\uparrow$ \\
\midrule
Lumina-DiMOO \tnote{a} & 0.99 & 0.93 & 0.85 & 0.84 & 0.84 & 0.71 & 0.86 \\
Lumina-DiMOO (ReMDM) & \textbf{1.00} & \textbf{0.94} & 0.86 & 0.87 & 0.82 & 0.74 & 0.87 \\
Lumina-DiMOO (Ours) & 0.99 & \textbf{0.94} & \textbf{0.88} & \textbf{0.93} & \textbf{0.87} & \textbf{0.79} & \textbf{0.90} \\
\bottomrule
\end{tabular}
\begin{tablenotes}
\item[a] We reevaluate the pre-trained Lumina-DiMOO on the GenEval benchmark for fair comparison. The generation step is 64.
\end{tablenotes}
\end{threeparttable}
\end{table*}

\section{Proposed Method}

In this paper, we aim to investigate whether self-correction can be enabled in pre-trained multimodal masked diffusion models without any additional finetuning or retraining. To achieve training-free self-correction, we need to revisit the existing training-free framework proposed by \citep{Wang2025RemaskingDD}. They generalize the reverse process of the masked diffusion model from Eq.~\eqref{eq:reverse_process} to 
\begin{equation}\label{eq:remasking}
\begin{aligned}
q_\sigma\left(x_s | x_t, x\right) 
=
\begin{cases}
(1-\sigma_t) \delta_{x_t}(x_s) + \sigma_t \delta_{m}(x_s), & x_t \neq m, \\[1mm]
\frac{\alpha_s-(1-\sigma_t)\alpha_t}{1-\alpha_t} \delta_{x}(x_s) + \frac{1-\alpha_s-\sigma_t \alpha_t}{1-\alpha_t} \delta_m(x_s), & x_t = m.
\end{cases}    
\end{aligned}
\end{equation}
That is, once a token is generated, they still have the probability $\sigma_t$ to be remasked. 
\citep{Wang2025RemaskingDD} propose three remasking schedules of $\sigma_t$, and find the confidence-based schedule with rescale achieves the best performance. In this schedule, the predicted token probabilities at the time of generation are used as the criterion for remasking. Tokens with lower predicted probabilities are more likely to be remasked. The criterion is reasonable in the sense that the predicted token probabilities $p_\theta^d(x_t)$ are only valid when $x_t^d=m$. 

However, this criterion is limited in the sense that we need to evaluate the confidence of each position condition on current $x_t$ rather than previous time step, i.e., the time that the token is generated. Therefore, recent works \citep{Zhao2024InformedCF, Kim2025FineTuningMD} propose to reevaluate the predicted token probabilities for each generated position at each decoding step, i.e., $p(x_t^d)=p_{\theta}^d(x_t^d|x_t^{\backslash d})$ where $x_t^{\backslash d}$ denotes changing $d$-th dimension of $x_t$ to $m$ and $p_{\theta}^d(x_t^d|\cdot)$ denotes the predicted probability of token $x_t^d$ by the masked diffusion model. \citep{Zhao2024InformedCF} propose to train a hollow-transformer for efficient estimation, while \citep{Kim2025FineTuningMD} propose to finetune the pre-trained masked diffusion model to estimate the predicted token probability. A natural question raised: 
\textit{Given the inductive bias of the pre-trained masked diffusion model, can we achieve training-free estimation for the predicted token probability?}

\subsection{Rationality of Training-free Self-Correction}
In this subsection, we use the real experimental results to verify the inductive bias of a pre-trained masked diffusion model. We utilize a state-of-art multimodal masked diffusion model \citep{Xin2025LuminaDiMOOAO}. We aim to investigate whether a pre-trained network is capable of identifying incorrect tokens during the generation process. Specifically, we sample multiple image-text prompt pairs $(z_0^{\text{image}}, z^{\text{prompt}})$ from the Flickr8k dataset~\citep{hodosh2013framing}. Using a pre-trained language and image tokenizer, we project $(z_0^{\text{image}}, z^{\text{prompt}})$ into the embedding space $(x_0, y)$. We then apply the noising process of masked diffusion models, $q_{t|0}(x_t | x_0)$, to obtain a noisy representation $x_t$, where $t \sim U(0,1)$ and $U(0,1)$ denotes the uniform distribution on the unit interval $(0,1)$. Next, we randomly select unmasked tokens and artificially flip their values. These manipulated tokens are fed into the pre-trained network, and we evaluate whether the model can detect the flipped tokens by examining the likelihood assigned to them. 

\begin{figure}[t]
    \centering
    \includegraphics[width=0.8\linewidth]{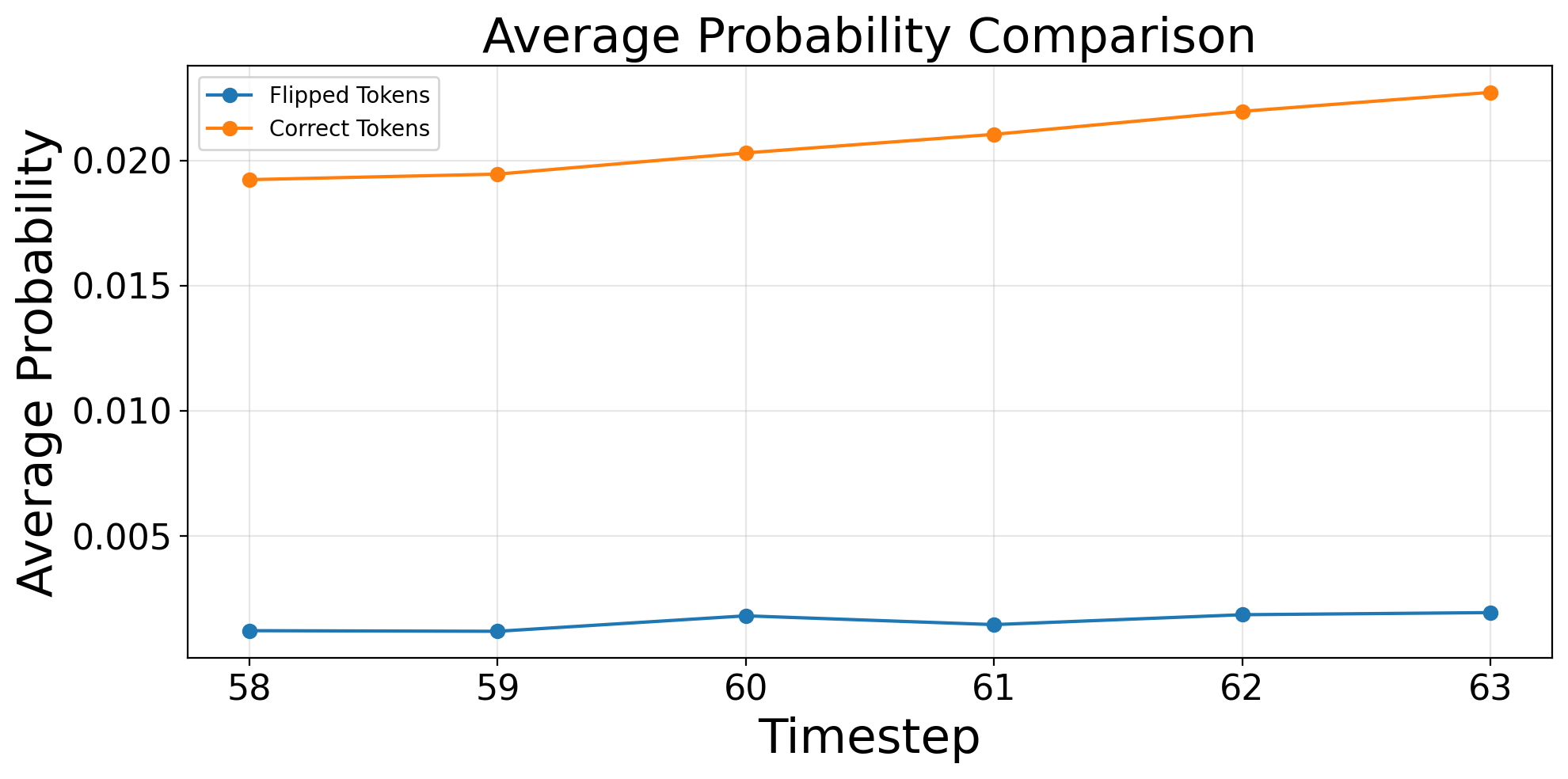}
    \caption{Average predicted probability of flipped tokens and correct tokens over 2000 samples. The x-axis denotes the time steps for generation (64 steps in total for text-to-image generation), while the y-axis denotes the average probability over all flipped positions and the correct position.}
    \label{fig:flip}
\end{figure}

We plot the average likelihood values of the original tokens and their flipped counterparts across different time steps in Figure~\ref{fig:flip}. The results reveal that the pre-trained masked diffusion model will assign significantly higher predicted probabilities to the tokens that make more sense (corrected tokens) than the tokens that does not make sense (flipped tokens). This phenomenon reflected the inductive bias of the pre-trained model on the tokens that are already generated. 

\begin{figure}[t]
    \centering
    \includegraphics[width=0.8\linewidth]{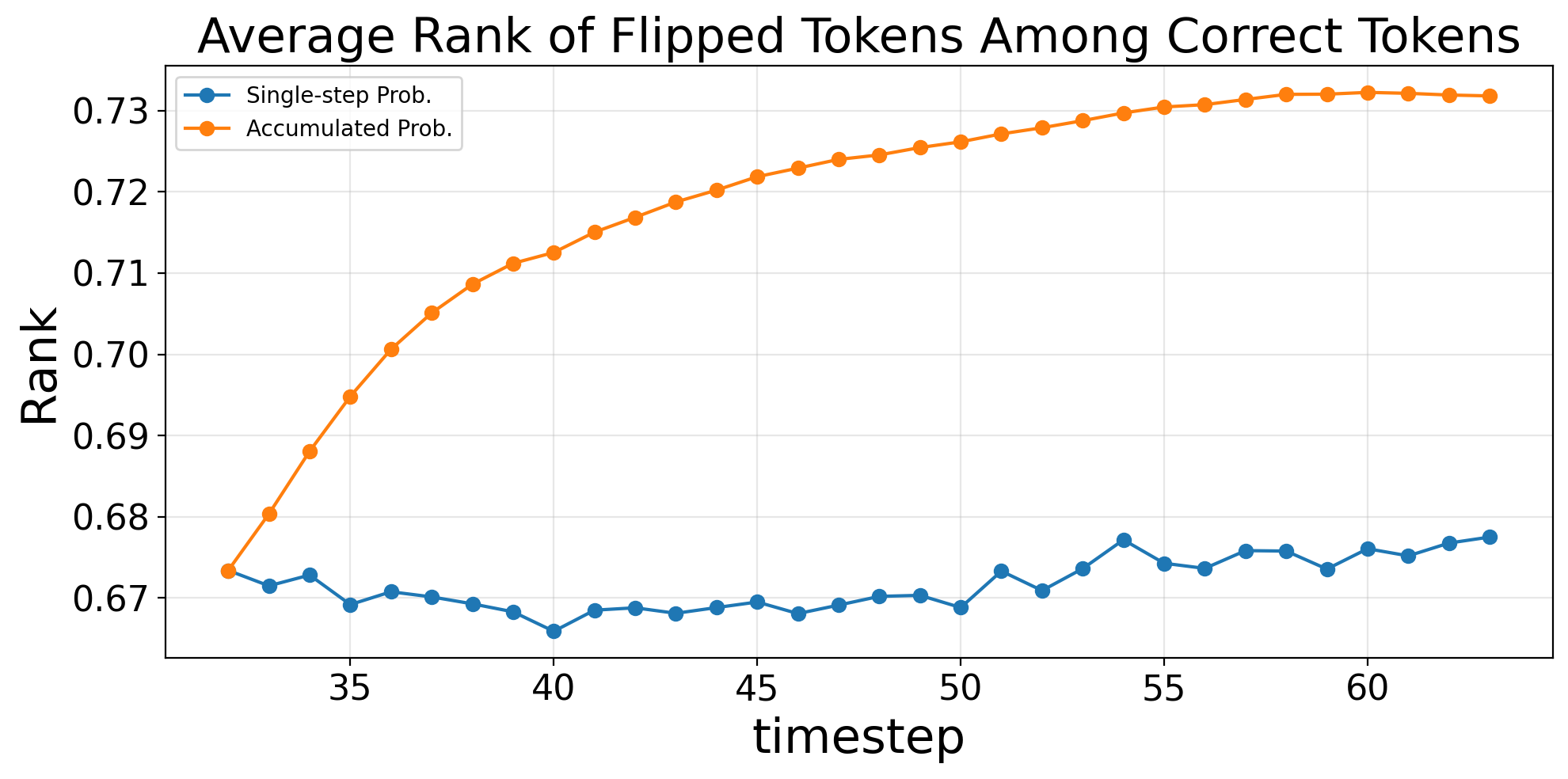}
    \caption{The effectiveness of using accumulated predicted probability. The x-axis denotes the time steps for generation, while the y-axis denotes the average rank of the predicted probabilities of flipped tokens among correct tokens. The larger the rank is, the smaller the probability is.}
    \label{fig:accumulated}
\end{figure}

In Fig \ref{fig:accumulated}, we demonstrate that using the accumulated predicted probabilities could distinguish the wrong tokens well. For each time step after timestep 32, we found the average rank of accumulated probability for the wrong tokens is more and more greater than the average rank evaluated on a single step. Therefore, it inspires us to design a sliding window for self-correction.


\begin{algorithm}
\caption{Training-free Self-Correction for Masked Diffusion Models}
\label{alg:remdm_sampling_alg}
\begin{algorithmic}
\REQUIRE Pre-trained denoising network $\mathbf{x}_\theta$, number of steps $T$, noise schedule $\{\alpha_t\}$, remasking schedule $\{\sigma_t\}$, remasking score type $\mathsf{Score}\in\{\text{Current},\text{Cumulated}\}$, remasking rule $\mathsf{Rule}\in\{\text{Deterministic},\text{Stochastic}\}$
\ENSURE Sample $\mathbf{x}_0$

\STATE Initialize $\mathbf{x}_T \gets \boldsymbol{m}$
\STATE Initialize per-token score $\mathbf{S} \gets \mathbf{0}$

\FOR{$i = T, T-1, \ldots, 1$}
    \STATE $t \gets i/T$, \quad $s \gets (i-1)/T$
    \STATE Retrieve $\alpha_t, \alpha_s, \sigma_t$

    \STATE \textbf{(1) Standard reverse update}
    \STATE $\hat{\mathbf{x}} \gets \mathbf{x}_\theta(\mathbf{x}_t)$
    \STATE Sample $\mathbf{x}_s \sim q_\sigma\!\left(\mathbf{x}_s \mid \mathbf{x}_t, \hat{\mathbf{x}}\right)$ \COMMENT{in Eq \eqref{eq:remasking}}
    \STATE $\mathbf{x}_t \gets \mathbf{x}_s$

    \STATE \textbf{(2) Training-free likelihood evaluation}
    \FOR{each position $d \in [\mathcal D]$ with $x_t^d \neq m$}
        \STATE $\ell_t^d \gets p_\theta^d\!\left(x_t^d \mid \mathbf{x}_t^{\backslash d}\right)$ \COMMENT{reevaluate likelihood}
        \IF{$\mathsf{Score} = \text{Current}$}
            \STATE $S^d \gets \ell_t^d$
        \ELSE
            \STATE $S^d \gets S^d + \log(\ell_t^d)$ \COMMENT{cumulated likelihood}
        \ENDIF
    \ENDFOR

    \STATE \textbf{(3) Remasking step}
    \STATE $K_t \gets \lfloor \sigma_t \cdot D \rfloor$
    \IF{$\mathsf{Rule} = \text{Deterministic}$}
        \STATE $\mathcal{R} \gets \operatorname{arg\,min}_{|\mathcal{R}|=K_t} \{S^d\}_{d: x_t^d \neq m}$ \COMMENT{remask $K_t$ lowest-score tokens}
    \ELSE
        \STATE Sample $\mathcal{R}$ of size $K_t$ with probability $\propto \exp(-S^d)$ over $\{d: x_t^d \neq m\}$
    \ENDIF
    \FOR{each $d \in \mathcal{R}$}
        \STATE $x_t^d \gets m$
    \ENDFOR
\ENDFOR

\STATE \textbf{return} $\mathbf{x}_0$
\end{algorithmic}
\end{algorithm}

\subsection{In-depth Design of Self-Correction Criteria}\label{sec:design}
Considering the inductive bias of the pre-trained masked diffusion model, we propose to use the predicted probability $p_\theta^d(x_t)$ for remasking, even when $x_t$ has already been generated. This design allows the model to fully exploit its capabilities throughout the generation process. 

Given this formulation, multiple design choices arise for the remasking strategy. We summarize several key aspects below:

\begin{itemize}
    \item \textbf{Remasking rule}: Remasking can be performed either by applying a hard rule (e.g., a Top-$K$ strategy analogous to that used during generation) or by sampling from a multinomial distribution defined by the predicted probability $p_\theta^d(x_t)$.
    \item \textbf{Temporal aggregation:}
    The remasking decision may depend on an accumulation of the predicted probabilities $p_\theta^d(x_t)$ across generation steps.
    \item \textbf{Top-$K$ margin for remasking:} Whether Top-K margin strategy used in generation can also provide an improvement on remasking 
    \item \textbf{Distributional uncertainty-based remasking}: Inspired by recent work \citep{Kim2025KLASSKF}, we consider utilizing richer information from the full vocabulary distribution of the predicted probability $p_\theta^d(x_t)$, rather than only relying on Top-$K$ or Top-$K$ margin. Specifically, we use either the KL divergence or the Wasserstein distance between the predicted distributions at consecutive time steps as a criterion to measure the model’s confidence at each position. Positions whose generated tokens exhibit larger distributional changes are considered less confident.
Accordingly, we remask the Top-$K$ positions with the largest distances during sampling.
\end{itemize}

We summarize the overall procedure in Algorithm~\ref{alg:remdm_sampling_alg}. In the following section, we carefully demonstrate the superiority of the proposed framework and provide a detailed comparison between various choices of remasking strategy.

\section{Experiments}\label{sec:exp}

\paragraph{Experimental Setup}
We conduct experiments using state-of-the-art multimodal masked diffusion models \citep{Xin2025LuminaDiMOOAO} and apply our self-correction strategy purely at inference time, without any additional fine-tuning or retraining. To assess the effectiveness of the proposed remasking strategy, we evaluate its performance on both text-to-image generation and multimodal understanding tasks. For text-to-image generation, we adopt the widely used GenEval benchmark \citep{Ghosh2023GenEvalAO}, which focuses on complex compositional prompts involving object counting, spatial relations, and attribute binding. For multimodal understanding, we report results on MMBench (MMB) \citep{Liu2023MMBenchIY}, SEED-Bench \citep{Li2023SEEDBenchBM}, and MMMU \citep{Yue2023MMMUAM}. For all multimodal understanding benchmarks, we set the maximum generation length to 256 tokens per question. To obtain an automatic and consistent correctness decision, we adopt GPT-4o-mini as the verifier: given the model response and the benchmark-provided reference, the verifier outputs the judgment. Unless otherwise specified, we use the same backbone, tokenizers, and evaluation protocols as the original model and only change the sampling procedure by inserting the proposed remasking step.

\begin{figure*}
    \centering
    \includegraphics[width=0.95\linewidth]{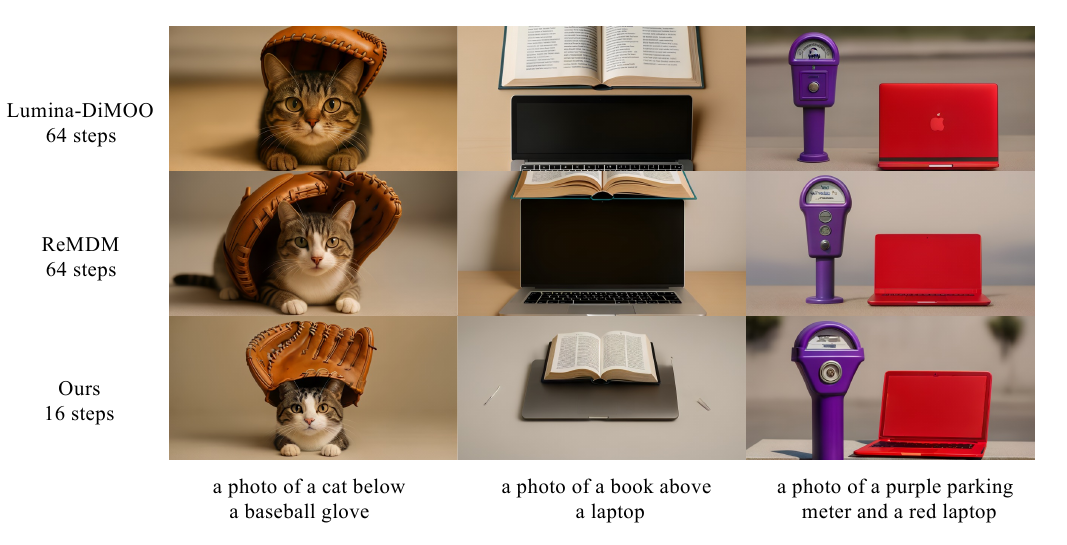}
    \caption{Comparison of generated images of Lumina-DiMOO, ReMDM, and our methods. Our method achieves better quality with less sampling steps.}
    \label{fig:geneval_demo}
\end{figure*}

\subsection{Overall Effectiveness}
\begin{figure}[t!]
    \centering
    \includegraphics[width=0.6\linewidth]{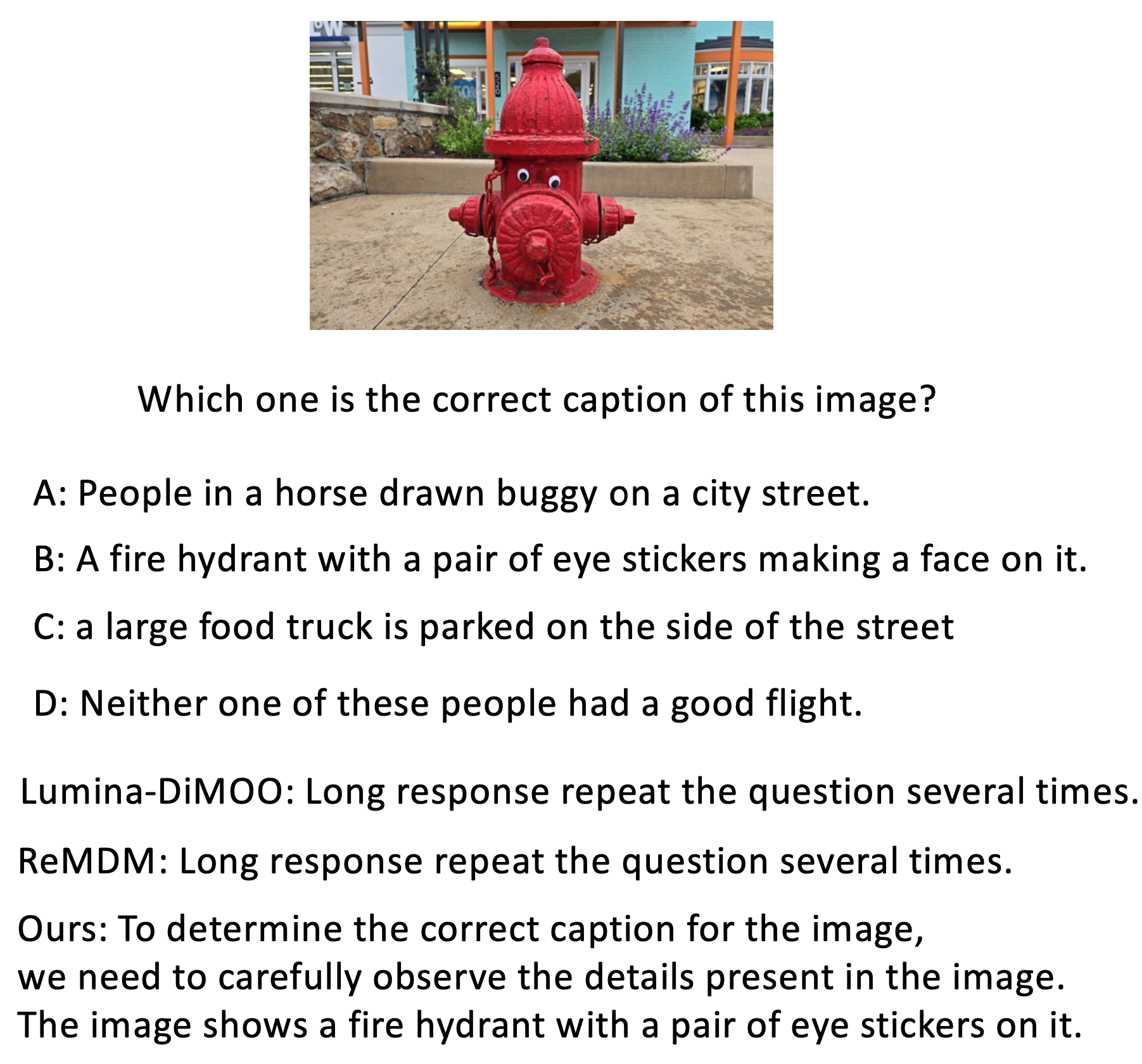}
    \caption{Illustration of the performance on multimodal understanding.}
    \label{fig:Illustration_mmbench}
\end{figure}

We first evaluate the overall effectiveness of the proposed remasking strategy. As shown in Table~\ref{tab:geneval}, our method improves GenEval performance over vanilla Lumina-DiMOO as well as existing training-free self-correction approaches~\citep{Wang2025RemaskingDD}. On multimodal understanding shown in Table~\ref{tab:multimodal} and Fig \ref{fig:Illustration_mmbench}, we observe consistent gains when enabling our remasking-based self-correction at inference time. Overall, these results support our key claim: the inductive bias of pre-trained masked diffusion models can be exploited to identify and revise low-confidence tokens, leading to improved final generation quality without any additional training.

\begin{table*}[tb]
\caption{Evaluation of multimodal understanding on several benchmarks. 
}
\label{tab:multimodal}
\centering

\begin{threeparttable}
\setlength{\tabcolsep}{8pt} 
\begin{tabular}{l c c c}
\toprule
Model & MMB $\uparrow$ & SEED $\uparrow$ & MMMU $\uparrow$ \\
\midrule
Lumina-DiMOO \tnote{a} & 58.7 & 71.4 & 41.4 \\
Lumina-DiMOO (ReMDM)  & 57.8 & \textbf{74.3} & 43.4 \\
Lumina-DiMOO (Ours)  & \textbf{60.7} & 74.2 & \textbf{44.0} \\
\bottomrule
\end{tabular}
\begin{tablenotes}
\item[a] We reevaluate the pre-trained Lumina-DiMOO on several multimodal understanding benchmarks for fair comparison. The generation step is 64 and the generated sequence length is 256.
\end{tablenotes}
\end{threeparttable}
\end{table*}

\subsection{Ablation Study on Remasking Design Choices}
We perform a comprehensive ablation study to analyze the impact of different remasking design choices described in Section~\ref{sec:design}.
Specifically, we compare various remasking rules, temporal aggregation strategies, and uncertainty criteria to better understand their individual contributions in Table \ref{tab:geneval_abl}. In general, ``Cumulated likelihood + deterministic remasking'' achieves the best performance. It verifies the observation that accumulated predicted probabilities help in identifying the wrong token shown in Fig \ref{fig:accumulated}. It also verifies the effectiveness of the widely used Top-$K$ strategy. We conduct a careful hyperparameter search for the performance shown in Table \ref{tab:geneval_abl}. The hyperparameter ``remasking schedule'' $\sigma_t$ is linked to the number of tokens $K_t$ to be remasked by $K_t= \lfloor \sigma_t \cdot D \rfloor$, where $\lfloor \cdot \rfloor$ denotes the floor function. We determined the hyperparameter through the performance on the validation set.

\begin{table*}[tb]
\centering
\caption{Ablation study on various remasking strategy designs on the GenEval benchmark. ``Deterministic remasking'' stands for a greedy algorithm that remasks the positions where the predicted probabilities are the smallest $K$ positions, while ``stochastic remasking'' represents the algorithm that samples the positions to be remasked according to the predicted probabilities. The smaller the predicted probability is, the more likely the position will be remasked.}
\label{tab:geneval_abl}
\resizebox{\textwidth}{!}{%
\begin{tabular}{l c c c c c c c}
\toprule
Method & Single & Two & Count & Color & Pos. & Attr. & Overall $\uparrow$ \\
\midrule
Lumina-DiMOO & 0.99 & 0.93 & 0.85 & 0.84 & 0.84 & 0.71 & 0.86 \\
\quad+ Current step likelihood + deterministic remasking & 0.95 & 0.92 & 0.83 & 0.87 & 0.84 & 0.76 & 0.86 \\
\quad+ Current step likelihood + stochastic remasking & 0.96 & 0.94 & 0.86 & 0.86 & 0.84 & 0.75 & 0.87 \\
\quad+ Cumulated likelihood + deterministic remasking & 0.99 & 0.94 & 0.88 & \textbf{0.93} & \textbf{0.87} & \textbf{0.79} & \textbf{0.90} \\
\quad+ Cumulated likelihood + stochastic remasking & 0.99 & 0.93 & 0.85 & 0.88 & 0.82 & 0.78 & 0.87 \\
\midrule
\quad+ Top K margin & 0.99 & 0.93 & \textbf{0.90} & 0.89 & 0.86 & 0.75 & 0.89 \\

\quad+ KL divergence & 0.98 & \textbf{0.95} & 0.85 & 0.86 & 0.82 & 0.72 & 0.86 \\

\quad+ Wasserstein distance & \textbf{1.00} & 0.93 & 0.84 & 0.88 & 0.84 & 0.66 & 0.86 \\
\bottomrule
\end{tabular}}
\end{table*}

\subsection{Remasking for Efficient Sampling}
\begin{figure}[t]
    \centering
    \includegraphics[width=0.8\linewidth]{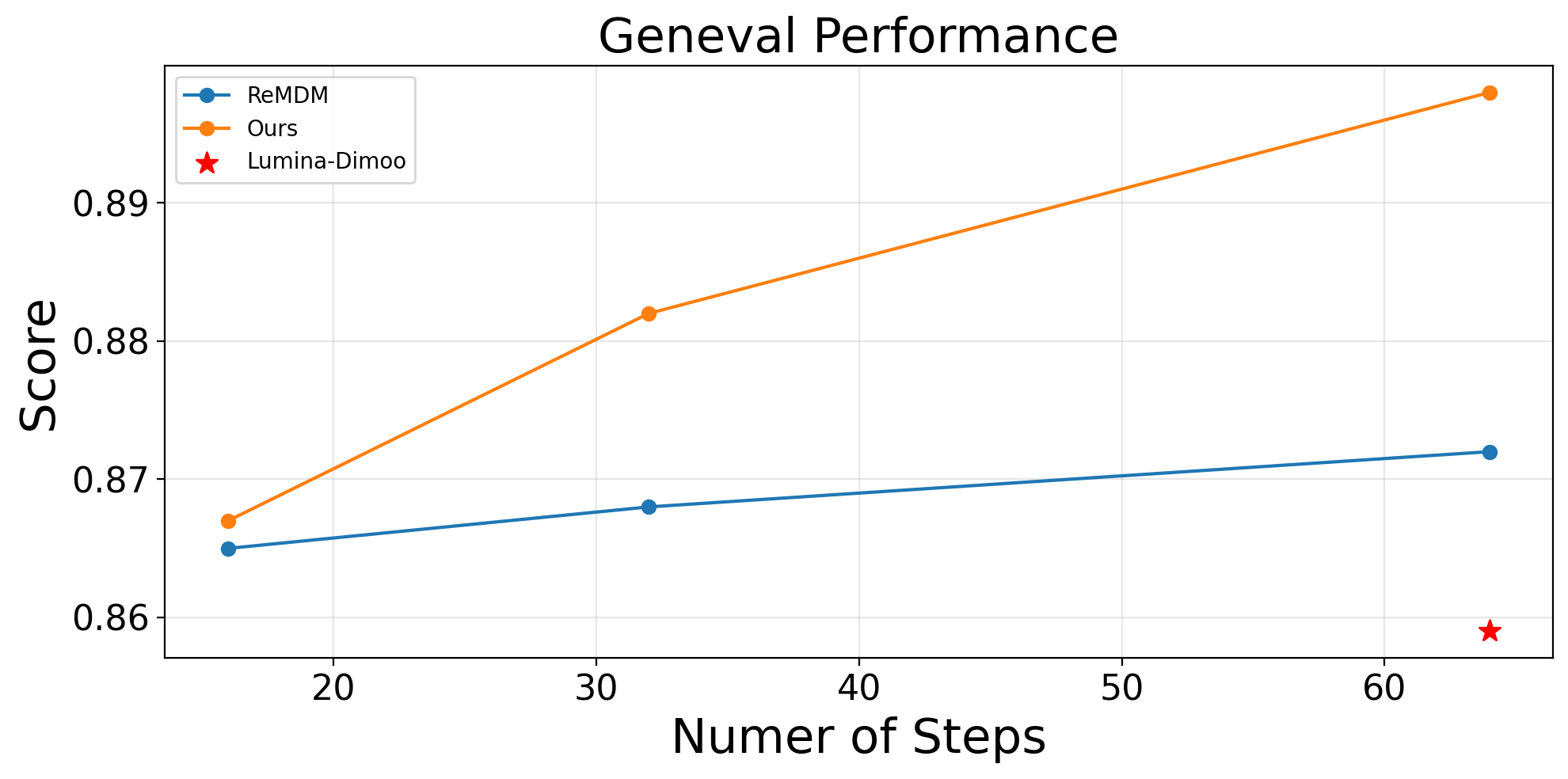}
    \caption{GenEval performance under different inference steps. The x-axis denotes the number of steps for generation, while the y-axis denotes the overall performance on the GenEval benchmark.}
    \label{fig:geneval-efficient}
\end{figure}

We next study whether the proposed remasking mechanism can serve as a practical acceleration technique at inference time.
We vary the number of reverse steps while keeping all other settings fixed, and report the GenEval overall score.
As shown in Figure~\ref{fig:geneval-efficient}, with only 16 sampling steps, our method achieves comparable or better GenEval performance than the 64-step Lumina-DiMOO baseline.
These results indicate that selectively revisiting low-confidence positions can mitigate early-step mistakes and enable four times speed up without sacrificing generation quality. The proposed method achieves more improvement against ReMDM, demonstrating the effectiveness of utilizing the inductive bias of the pre-trained model. We also provide some visualization of the generated images in Fig \ref{fig:geneval_demo}.

\begin{table*}[tb]
\centering
\caption{Evaluation of the proposed method on the GenEval benchmark based on MMaDA-8B-MixCoT \citep{yang2025mmada}.}
\label{tab:geneval_mmada}
\begin{threeparttable}
\resizebox{\textwidth}{!}{%
\begin{tabular}{l c c c c c c c}
\toprule
Method & Single & Two & Count & Color & Pos. & Attr. & Overall $\uparrow$ \\
\midrule
MMaDA-8B-MixCoT \footnote{a} (Top K)          
& \textbf{0.93} & 0.47 & 0.31 & 0.81 & 0.16 & 0.26 & 0.49 \\
MMaDA-8B-MixCoT (Top K margin)   
& 0.88 & 0.49 & 0.29 & 0.79 & 0.15 & \textbf{0.28} & 0.48 \\
MMaDA-8B-MixCoT (ReMDM) 
& 0.91 & 0.59 & 0.38 & 0.78 &\textbf{0.20} & 0.19 & 0.51 \\
MMaDA-8B-MixCoT (Ours)           
& 0.91 & \textbf{0.67} & \textbf{0.39} & \textbf{0.84} & 0.15 & 0.18 & \textbf{0.52} \\
\bottomrule
\end{tabular}}
\end{threeparttable}
\begin{tablenotes}
\item[a] We reevaluate the pre-trained MMaDA-8B-MixCoT on GenEval benchmarks for fair comparison. The generation step is 64.
\end{tablenotes}
\end{table*}
\subsection{Generalization Across Backbone Models}
Finally, we evaluate whether our remasking strategy is tied to a specific backbone or can generalize across masked diffusion architectures. In Table \ref{tab:geneval_mmada}, 
we evaluate the performance of the proposed method on MMaDA-8B-MixCoT \citep{yang2025mmada} for text-to-image generation. We observe consistent improvements when enabling the same inference-time self-correction rule.
This suggests that our method primarily leverages generic inductive biases of masked diffusion models, highlighting the robustness of our approach. 

\vspace{-10pt}

\section{Conclusion}
In this work, we provided an explanation for the error accumulation phenomenon caused by parallel and irreversible token updates. Then, we proposed a model-agnostic training-free self-correction framework that operates directly at inference time. Unlike prior approaches that rely on additional training, auxiliary evaluators, or misaligned likelihood, our method leverages the inductive biases of pre-trained masked diffusion networks for self-correction. Extensive experiments on text-to-image generation and multimodal understanding benchmarks demonstrate that our approach consistently improves generation quality, while also enabling faster sampling without sacrificing output quality. The observed gains generalize across different masked diffusion architectures, highlighting the robustness and practical applicability of the proposed method.
\bibliography{references}
\bibliographystyle{plain}

\appendix
\onecolumn
\clearpage

\end{document}